\documentclass[conference]{IEEEtran}
\IEEEoverridecommandlockouts
\usepackage{cite}

\usepackage{caption}
\usepackage{subcaption}

\usepackage{amsmath,amssymb,amsfonts}
\usepackage{algorithmic}
\usepackage{graphicx}
\usepackage{textcomp}
\usepackage{xcolor}
\def\BibTeX{{\rm B\kern-.05em{\sc i\kern-.025em b}\kern-.08em
    T\kern-.1667em\lower.7ex\hbox{E}\kern-.125emX}}
\begin{document}

\title{Step Counting with Attention-based LSTM}

\author{\IEEEauthorblockN{Shehroz S. Khan and Ali Abedi}
\IEEEauthorblockA{\textit{KITE Research Institute, Toronto Rehabilitation Institute} \\
\textit{University Health Network}\\
Toronto, Canada \\
\{shehroz.khan, ali.abedi\}@uhn.ca}
}

\maketitle

\begin{abstract}
Physical activity is recognized as an essential component of overall health. One measure of physical activity, the step count, is well known as a predictor of long-term morbidity and mortality. Step Counting (SC) is the automated counting of the number of steps an individual takes over a specified period of time and space. Due to the ubiquity of smartphones and smartwatches, most current SC approaches rely on the built-in accelerometer sensors on these devices. The sensor signals are analyzed as multivariate time series, and the number of steps is calculated through a variety of approaches, such as time-domain, frequency-domain, machine-learning, and deep-learning approaches. Most of the existing approaches rely on dividing the input signal into windows, detecting steps in each window, and summing the detected steps. However, these approaches require the determination of multiple parameters, including the window size. Furthermore, most of the existing deep-learning SC approaches require ground-truth labels for every single step, which can be arduous and time-consuming to annotate. To circumvent these requirements, we present a novel SC approach utilizing many-to-one attention-based LSTM. With the proposed LSTM network, SC is solved as a regression problem, taking the entire sensor signal as input and the step count as the output. The analysis shows that the attention-based LSTM automatically learned the pattern of steps even in the absence of ground-truth labels. The experimental results on three publicly available SC datasets demonstrate that the proposed method successfully counts the number of steps with low values of mean absolute error and high values of SC accuracy.
\end{abstract}


\section{Introduction}
\label{sec:introduction}
As the fundamental unit of human locomotion, steps are the preferred method of quantifying ambulatory physical activity \cite{bassett2017}. The association between steps per specific time periods and health variables has been reported in several cross-sectional studies \cite{pillay2015,paluch2022,cuthbertson2022}. To illustrate, a higher number of steps is inversely associated with the risk of cardiovascular events and premature death \cite{sheng2021}. Step Counting (SC) is the automated counting of the number of steps an individual takes over a specified period of time and space. SC has applications for telemonitoring/telemedicine to measure the number of steps and monitor the daily physical activity of patients remotely \cite{thorup2016}. There are many other applications for SC, including indoor navigation where global positioning systems are unreliable, and pedestrian dead reckoning \cite{flores2018}.

\begin{figure}[]
\centerline{\includegraphics[scale=.275]{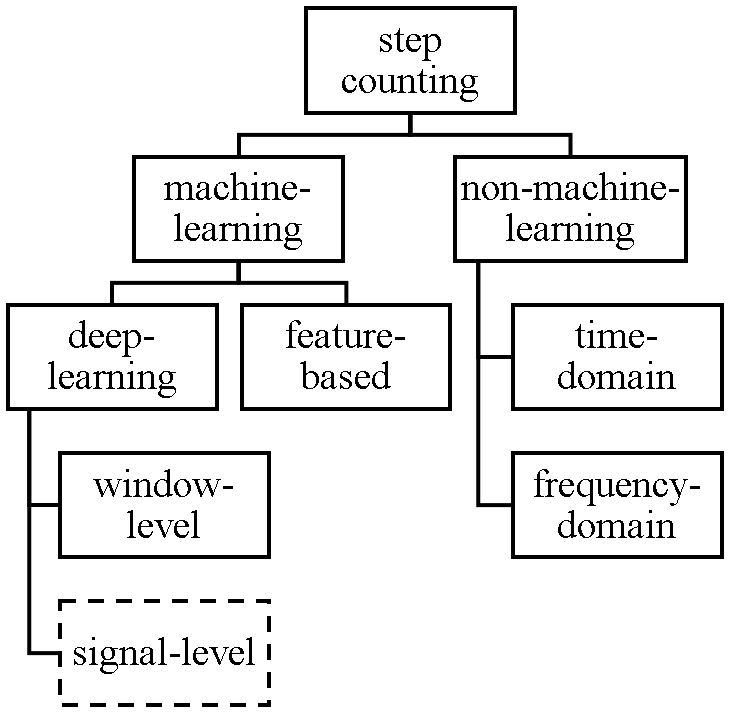}}
\caption{Hierarchical classification of the existing SC approaches. The proposed method in this paper, dashed box, is the only deep-learning-based signal-level SC method that analyzes the entire signal as a whole and solves the SC as a regression problem (see Sections \ref{sec:introduction}, \ref{sec:related_work},  and \ref{sec:method}).}
\label{fig:methods}
\end{figure}

The increasing ubiquity of smartphones and smartwatches equipped with a variety of built-in Inertial Measurement Unit (IMU) sensors, such as accelerometers, gyroscopes, and magnetometers, has led to the development of various SC methods using the multivariate time series of sensor signals. The existing SC approaches are broadly categorized into non-machine-learning- and machine-learning-based approaches. The non-machine-learning-based approaches can be divided into time-domain and frequency-domain approaches, Fig. \ref{fig:methods}.

Time-domain approaches generally rely on thresholding or peak detection. In thresholding methods, a step is detected when sensor data satisfy predefined criteria. These methods are particularly effective when detecting movements at the foot, where heel strikes can cause large and short-lived accelerations \cite{scholkmann2012}. Peak detection or zero-crossing methods usually work on low-pass filtered signals and detect the occurrence of steps according to the presence of peaks in the signal. Based on the peaks and the distance between the peaks, some methods try to find the inherent periods in the signal. Auto-correlation is another method to detect the period of signal and step accordingly \cite{jayalath2013}. Some methods work based on stride in which the stride template is formed offline and cross-correlated with the signal \cite{brajdic2013}. Frequency-domain approaches, on the other hand, generally use the Fourier transforms of the signals, such as short-term Fourier transform and wavelet transform, and utilize the features in the frequency domain to detect steps \cite{brajdic2013}.

A major limitation of non-machine-learning-based approaches for SC is that they require careful tuning of several parameters \cite{flores2018,brajdic2013}. For instance, in thresholding, peak detection, and period detection approaches in the time domain, the main difficulty is finding an optimal threshold/criteria to detect a specific timestep of the signal as a peak or consider segments of the signal as a period. Optimizing the window length for short-term Fourier transform and the parameters of continuous or discrete wavelet transforms is the issue with frequency-domain approaches \cite{brajdic2013}.

The machine-learning-based SC approaches are categorized into feature-based and deep-learning approaches. In the former, features such as mean, variance, standard deviation, energy, and entropy are first extracted from the windows of signal in the time/frequency domain and then classified into step versus non-step using traditional machine-learning techniques such as support vector machines and Hidden Markov Model (HMM) trained on sequences of features \cite{mannini2011}. The traditional machine-learning-based approaches also suffer from the need for parameter tuning, e.g., the selection of most effective features and the length of windows in which the features are extracted \cite{brajdic2013,flores2018}. In deep-learning methods, neural networks can learn features from raw sensor signals. The existing deep-learning-based SC methods are mostly based on Convolutional Neural Networks (CNNs) and Recurrent Neural Networks (RNNs) \cite{luu2022,chen2018,bagui2022,pillai2020,ren2021,shao2018}.

In most of the above-discussed approaches, the steps are first detected by thresholds, features, CNNs, or RNNs, and then the total number of steps is output \cite{luu2022,chen2018,bagui2022,pillai2020,ren2021,shao2018}. One practical problem with such a setup is that to evaluate the algorithms, ground truth is needed for every step taken. Manually labelling such fine-grained labelled data for each step is arduous or time-consuming. Alternatively, additional hardware may need to be added (e.g., pressure sensors) on the heels of the shoes, which is infeasible in a real-world setting. Keeping these issues in mind, in this paper, we propose a novel SC method that overcomes the limitations mentioned thus far, namely the necessity to determine the window size and the need for ground-truth data for single steps. Our main contributions are as follows:
\begin{itemize}
    \item For the first time, SC is formulated as a regression problem that is solved using an attention mechanism for many-to-one LSTMs capable of analyzing variable-length sequences.
    \item The SC problem is solved at the signal level that is capable of analyzing the entire input time-series signal as a whole.
    \item Extensive experiments conducted on three publicly available SC datasets \cite{brajdic2013,flores2018,mattfeld2017} show the superiority of our approach to a variety of machine-learning and non-machine-learning SC methods.
\end{itemize}

Our approach does not require windowing or annotation of individual steps - only the final count of steps is sufficient for training deep-learning models.

This paper is structured as follows. In Section \ref{sec:related_work}, we briefly study the existing deep-learning SC approaches. Section \ref{sec:method}, introduces the proposed method for SC. Section \ref{sec:experiments} describes experimental settings and results on the proposed methodology. In the end, Section \ref{sec:conclusions} presents our conclusions and directions for future works.

\section{Related Work}
\label{sec:related_work}
In this section, we discuss some of the deep learning methods for SC from IMU sensors. Unfortunately, there are not many papers published on SC using deep learning. There are, however, some preprints available. Shao et al. \cite{shao2018} proposed a method for SC through step detection. A window is slid over the input signal, and a CNN classifies the signal in the window as a left step, right step, or no step. The CNN consists of a 1-dimensional convolutional layer followed by fully-connected layers. Chen \cite{chen2018} proposed a SC method using a many-to-many Long Short-Term Memory (LSTM), having the sequences of windows extracted from the signal as input and outputting step vs. non-step for each window. In the method proposed by Pillai et al. \cite{pillai2020}, the input signal is first segmented into windows, and the signal in each window is analyzed by an LSTM-based neural network. The last timestep of the LSTM is trailed by fully-connected layers to output left step or right step classes. Luu et al. \cite{luu2022} extended the work by Pillai et al. \cite{pillai2020}, where instead of the last timestep of the LSTM, the output of the LSTM over all timesteps followed by fully-connected layers are used to output step or non-step classes. In addition to LSTM, Luu et al. \cite{luu2022} have used WaveNet \cite{oord2016} and a 1D CNN for SC through step detection.

Even though the above deep-learning-based methods do not require tuning parameters such as a threshold on peak values to be considered steps as with the time-domain methods, the window size parameter still needs to be specified manually in these methods. The window size needs to be adjusted for different datasets using different sensors, with different sampling frequencies and with different sensor placements on the body of individuals, with different walking styles, walking speeds, ages, and disabilities. To illustrate, the window length used by Shao et al. \cite{shao2018} is "length of training step segments", Pillai et al. \cite{pillai2020} is 0.46 seconds "based on tuning experiments and literature", and Luu et al. \cite{luu2022} is 2 seconds, and 4 seconds for LSTM, and CNN, respectively. None of the above works have examined the effect of different window sizes in different situations.

In the next section, we describe our approach for SC, which only needs a final count of the steps for training and evaluation of deep-learning models and does not need window size as a parameter to tune.

\section{Step Counting through Regression using Attention in LSTM}
\label{sec:method}
Input data to the proposed attention-based many-to-one LSTM architecture for SC is time series of accelerometer sensor signals of variable length. The neural network solves a regression problem and outputs the number of specific learned patterns in the input time series corresponding to the number of steps in the accelerometer sensor signal. The raw x, y, and z components of the accelerometer sensor signal or their $l2$ norm (as multivariate or univariate time series) can be input to the neural network. The input time series is normalized into the range of 0 to 1 by subtracting the minimum value of the signal and dividing it by the range of the signal. We used LSTM as the main component of the proposed neural network architecture. It can either be the vanilla RNN, Identity RNN, or Gated Recurrent Unit \cite{sherstinsky2020}. These RNNs are capable of handling sequences of variable lengths and situations in which there are sequences of variable lengths in successive mini-batches of epochs.

The shape of the input tensor as a mini-batch is $(N, L_i, input\_size), i = 0, 1, ..., N$ in which $N$ is the number of samples in the mini-batch, $L_i$ is the length (number of timesteps) of the $i$-th signal sample, and $input\_size$ is the dimensionality of the signal at each timestep. The LSTM has $num\_layers$ layers where $num\_layers = 2$ would mean stacking two LSTMs together to form a stacked LSTM, with the second LSTM taking in outputs of the first LSTM and computing the final results. The LSTM outputs a tensor of size $(N, L_i, hidden\_size), i = 0, 1, ..., N$. To make the proposed method capable of analyzing variable-length sequences, the output tensor of LSTM for different samples in the current mini-batch are fed to the attention mechanism sample-by-sample. The $(N, L_i, hidden\_size)$ output tensor of LSTM is divided into $N$ tensors $h_i$ of size $(L_i, hidden\_size), i = 0, 1, ..., N$. A linear layer of size $hidden\_size \times hidden\_size$ is used as an attention layer, takes the outputs of LSTM sample-by-sample $h_i, i = 0, 1, ..., N$, and outputs energies $e_i$ of the same size $(L_i, hidden\_size)$. $e_i$ then will be multiplied by the summation of $h_i$ over length dimension ($s_i$) to output a vector whose softmax generates weights $w_i$ of size $L_i$. $w_i$, as the result of training the attention layer, is multiplied by $h_i$ to output context $c_i$ of length $hidden\_size$.

\begin{figure}[t]
     \centering
     \begin{subfigure}[b]{1\linewidth}
         \centering
         \includegraphics[scale=.225]{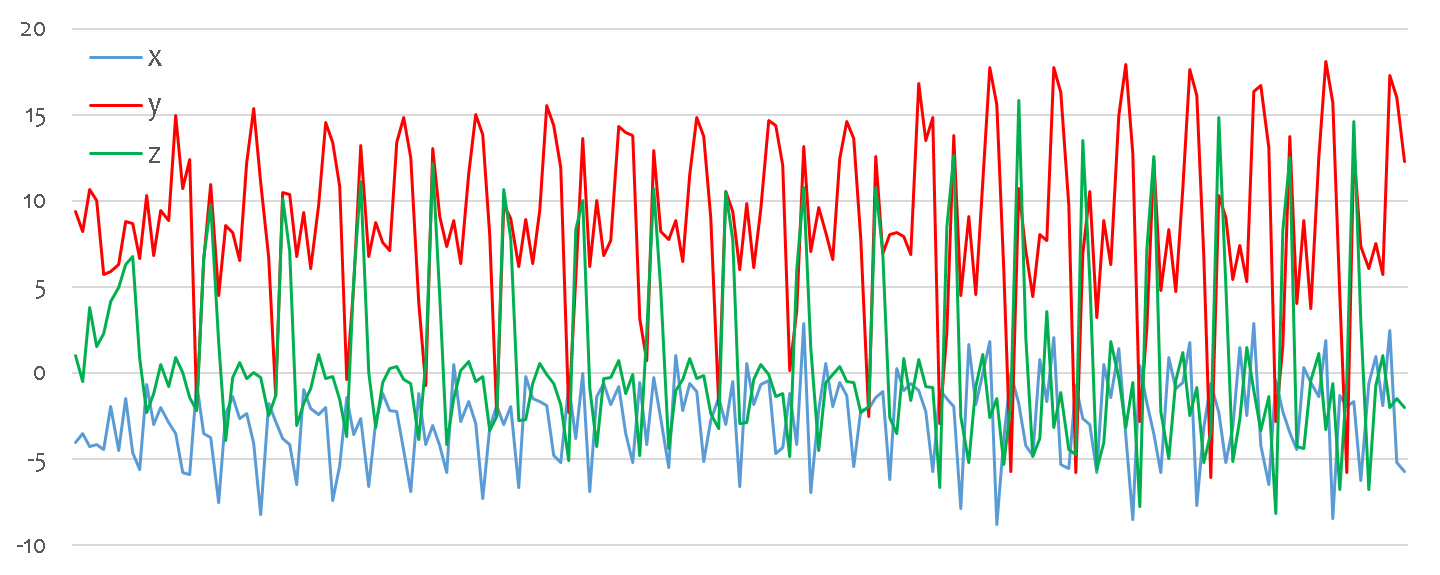}
         \caption{}
         \label{fig:exemplary_a}
     \end{subfigure}
     \hfill
     \centering
     \begin{subfigure}[b]{1\linewidth}
         \centering
         \includegraphics[scale=.225]{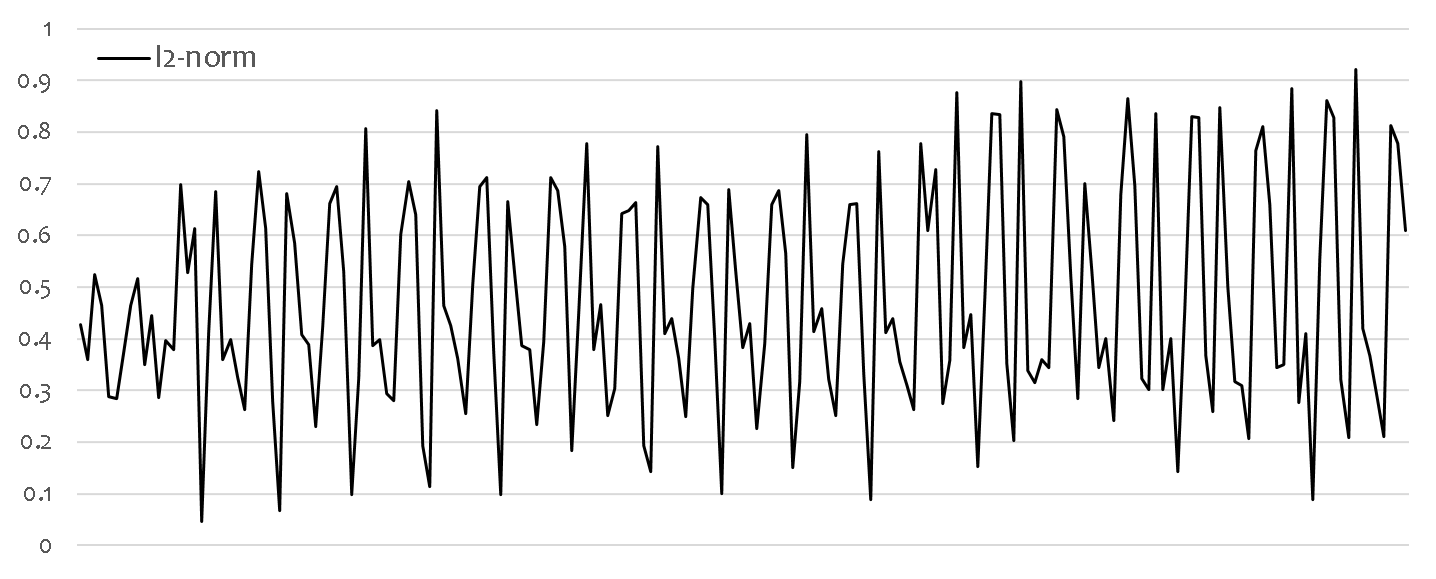}
         \caption{}
         \label{fig:exemplary_b}
     \end{subfigure}
     \hfill
          \centering
     \begin{subfigure}[b]{1\linewidth}
         \centering
         \includegraphics[scale=.225]{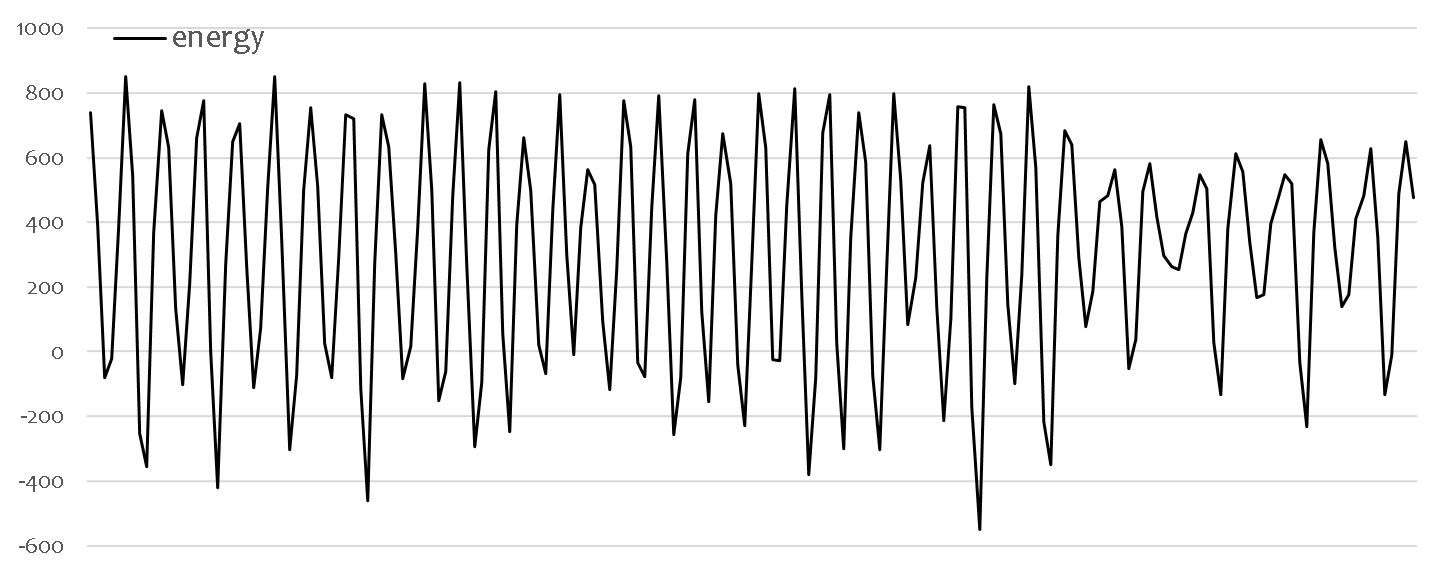}
         \caption{}
         \label{fig:exemplary_c}
     \end{subfigure}
     \hfill
\caption{(a) The raw $x$, $y$, and $z$ components, (b) the normalized $l2$-norm of (a), and (c) the energy (corresponding to the input signal in (b)) of the attention mechanism in a trained attention-based LSTM, for the first half of an exemplary accelerometer signal in the WDSC dataset.}
\label{fig:exemplary}
\end{figure}

The concatenations of contexts $c_i$ and summations $s_i$ for all the samples form a tensor of size $(N, hidden\_size \times 2)$. The concatenation tensor is fed to a linear layer of size ($hidden\_size \times 2) \times 1$ to generate the final output having a single real-valued number for each sample in the current mini-batch. The Mean Absolute Error (MAE) is used as the loss function of the neural network. During training, the attention layer and the corresponding weights (generated from the multiplication of the output of the attention layer and $s_i$) learn to pay attention to the specific patterns (steps) and their summation in the input time series (accelerometer sensor signal), see Fig. \ref{fig:exemplary} as an example.

The advantage of the proposed attention mechanism over the original versions of the attention mechanism for RNNs \cite{bahdanau2014,luong2015} is that it can handle variable-length sequences. We provide the input to the attention layer, followed by multiplications and concatenations, in a sample-by-sample manner for individual training samples (with different lengths) successively.


\section{Experiments}
\label{sec:experiments}
In this section, the performance of the proposed SC method is evaluated on three publicly available SC datasets using different evaluation metrics. We compare different settings of the proposed method with the previous machine-learning and non-machine-learning SC methods.

\subsection{Datasets}
\label{sec:datasets}
\noindent\textbf{WDSC:} Brajdic and Harle \cite{brajdic2013} collected and annotated a dataset for walk detection and SC. The data was collected from 27 subjects of different ages and heights using the built-in accelerometer of an Android smartphone sampled at 100 Hz and under six different smartphone placements. There are 117 samples (x, y, and z accelerometer signals of walks) in the dataset, each labelled based on the start and end time of the walk and the number of steps during the walk. The cropped signals from the start to end time of the walk are considered as the input, and the number of steps in the cropped signals is considered as the ground-truth labels. A major advantage of this dataset is the diversity of the location of the smartphone accelerometer sensors on the body. Table \ref{tab:dataset_statistics} presents the statistics of the samples in this dataset. Due to the lack of ground-truth labels for individual steps in the WDSC dataset \cite{brajdic2013}, the previous deep-learning SC approaches \cite{luu2022,chen2018,bagui2022,pillai2020,ren2021,shao2018} cannot be trained and evaluated on this dataset. For the first time, we show the results of a deep-learning SC approach on this dataset.

\begin{table}
\caption{The statistics, minimum, maximum, mean, standard deviation (STD), and skew of the ground-truth number of steps in the WDSC dataset \cite{brajdic2013}, WeAllWalk dataset \cite{flores2018}, and the regular and semi-regular parts of the Pedometer dataset \cite{mattfeld2017}.}
\centering
\begin{tabular}{
p{.225\linewidth}p{.1\linewidth}p{.1\linewidth}p{.15\linewidth}p{.175\linewidth}
}
\hline
 & WDSC \cite{brajdic2013}  & WeAll Walk \cite{flores2018} & Pedometer Regular \cite{mattfeld2017} & Pedometer Semi-regular \cite{mattfeld2017}\\ \hline
Minimum \char"0023steps & 63 & 2 & 857 & 548\\ \hline
Maximum \char"0023steps & 106 & 136 & 1100 & 814\\ \hline
Mean \char"0023steps & 78 & 40.71 & 991.03 & 704.03\\ \hline
STD \char"0023steps & 8.46 & 33.29 & 54.03 & 65.57\\ \hline
Skew \char"0023steps & 0.56 & 0.81 & -0.27 & -0.21\\ \hline
\end{tabular}
\label{tab:dataset_statistics}
\end{table}

\noindent\textbf{WeAllWalk:} Flores and Manduchi \cite{flores2018} collected and annotated a dataset from 15 subjects. The uniqueness of this dataset is due to the presence of sighted and blind subjects. Ten blind subjects (using a white cane or guide dog) and five sighted subjects contributed to the data collection. While walking ten different paths, the subjects carried two Apple iPhone 6S smartphones at two different body locations. Each smartphone recorded data from its accelerometer, gyroscope, and magnetometer at a rate of 25 Hz. The accelerometer signals are used in the experiments in this paper. There are 932 samples in the dataset categorized into three categories of 290 samples from sighted subjects, 468 samples from blind subjects with a white cane, and 174 samples from blind subjects with a guide dog. Table \ref{tab:dataset_statistics} presents the statistics of the samples in this dataset.

\noindent\textbf{Pedometer:} Mattfeld et al. \cite{mattfeld2017} collected and annotated a dataset from 30 subjects for SC. The subjects wore three Shimmer3 sensors on their non-dominant wrist, hip, and non-dominant ankle. Each sensor recorded accelerometer and gyroscope data at 15 Hz. Unique to this dataset is the availability of different walking regularities, including regular (walking a path), semi-regular (conducting a within-building activity), and unstructured (conducting a within-room activity). There are a total of 270 samples in the dataset, nine ankle-, hip-, and wrist-placement signals collected during regular, semi-regular, and unstructured walking from each of the 30 subjects. According to the authors' recommendations \cite{mattfeld2021}, 180 regular and semi-regular accelerometer signal samples of this dataset are used in our experiments, whose statistics are presented in Table \ref{tab:dataset_statistics}.

\subsection{Evaluation Metrics}
\label{sec:evaluation_metrics}
The researchers who collected the above datasets also developed various evaluation metrics for SC. Considering $\mathrm{y\_trues}$, and $\mathrm{y\_preds}$ as the ground-truth step count, and predicted step count, Brajdic and Harle \cite{brajdic2013} defined an Error Rate (ER) for each sample as follows. The mean and standard deviation of Equation \ref{eq:brajdic} over all test samples are reported in this paper.

\begin{equation}
\label{eq:brajdic}
\mathrm{
\frac{y\_preds - y\_trues}{y\_trues} \times 100
}
\end{equation}

Mattfeld et al. \cite{mattfeld2021} defined a Running Count Accuracy (RCA) for each sample (defined below) and considered the prediction as undercount, or overcount if the accuracy is less than, or greater than 1. The mean and standard deviation of Equation \ref{eq:mattfeld} over all test samples are reported in this paper.

\begin{equation}
\label{eq:mattfeld}
\mathrm{
\frac{y\_preds}{y\_trues}
}
\end{equation}

Flores and Manduchi \cite{flores2018} first calculated the number of samples with y\_preds $-$ y\_trues $<$ $0$ as undercount and the number of samples with y\_preds $-$ y\_trues $>$ $0$ as overcount, and defined their normalization as percentages of UnderCount (UC) and OverCount (OC) as follows.

\begin{equation}
\label{eq:flores_undercount}
\mathrm{
normalized\:undercount = \frac{undercount}{total\:number\:of\:samples} \times 100 
}
\end{equation}

\begin{equation}
\label{eq:flores_overcount}
\mathrm{
normalized\:overcount = \frac{overcount}{total\:number\:of\:samples} \times 100 
}
\end{equation}

Luu et al. \cite{luu2022} defined an ACCuracy (ACC) for each sample as follows. The mean and standard deviation of Equation \ref{eq:luu} over all test samples are reported in this paper.

\begin{equation}
\label{eq:luu}
\mathrm{
(1 - \frac{|y\_preds - y\_trues|}{y\_trues}) \times 100.
}
\end{equation}

Since the proposed SC method solves a regression problem, the MAE between $\mathrm{y\_trues}$ and $\mathrm{y\_preds}$ is also reported.

\subsection{Experimental Settings}
\label{sec:experimental_settings}
We implemented LSTM with and without attention mechanism to evaluate their performance. The proposed method is implemented with the 3-dimensional time series (x, y, and z components of the accelerometer signal) $input\_size = 3$ or 1-dimensional time series ($l2$-norm of the x, y, and z components) $input\_size = 1$ or their combination $input\_size = 4$. Compared to the previous methods, one major advantage of the proposed method is that it does not require many parameters that need to be changed depending on the dataset or signal. However, LSTM-specific parameters may still need to be set, which are as follows. The LSTM is unidirectional with $num\_layers = 2$ and $hidden\_size = 128$. The attention layer is a linear layer of size $hidden\_size \times hidden\_size$. The final fully-connected layers, after the attention mechanism, contains two linear layers of size $(hidden\_size \times 2) \times hidden\_size$ and $hidden\_size \times 1$. For the LSTM with no attention, the architecture of the LSTM and the final fully-connected layers is the same as the attention-based network. Due to not improving the results, no dropout is used in both the LSTMs with and without attention.

As discussed in Section \ref{sec:datasets}, the frequency of WDSC, Pedometer, and WeAllWalk datasets are 100 Hz, 15 Hz, and 25 Hz. Before inputting the signals into the neural networks, the samples in the WDSC dataset are down-sampled by a factor of 4 to reduce the number of timestamps, and the samples in the Pedometer and WeAllWalk datasets remain unchanged. Then, the signals are normalized as explained in section \ref{sec:method}.

None of the datasets defined separate train and test sets. The non-machine-learning methods that used these datasets, e.g., \cite{mattfeld2021}, reported the results of their experiments on all the samples of the datasets. The machine-learning methods applied on these datasets, e.g., \cite{luu2022}, used cross-validation and reported the results of their algorithms on all the samples in the dataset over all the folds. In this paper, cross-validation is implemented for evaluation, and the results are reported for all the samples in the datasets. The architecture and hyper-parameters of the models are the same in all the folds of cross-validation in all the datasets.

The MAE is the loss function, and Adam is the optimizer. The batch size and the number of epochs are 16, and 250, respectively. The learning rate starts from 0.001 and is scheduled to be reduced by a factor of 10 in each 75 epochs. The experiments were implemented in PyTorch \cite{paszke2019} and scikit-learn \cite{pedregosa2011} on a server with 64 GB of RAM and NVIDIA TeslaP100 PCIe 12 GB GPU. The code of our implementations is available at https://github.com/abedicodes/stepcounting.

\begin{table*}[ht]
\caption{The results of the proposed SC method on the WDSC dataset \cite{brajdic2013} using the LSTM with and without attention. In LSTM$-a \times b$, $a$, and $b$ mean the number of layers in the LSTM, and the number of neurons in the layers, respectively. xyz means the raw $x$, $y$, and $z$ components of the accelerometer data and $l2$ means the $l2$-norm of xyz as the inputs to the neural networks, using different evaluation metrics (see Section \ref{sec:evaluation_metrics}).}
\centering
\begin{tabular}{
p{.3\linewidth}p{.075\linewidth}p{.075\linewidth}p{.085\linewidth}p{.085\linewidth}p{.085\linewidth}
}
\hline
 & MAE & UC, OC & ER & RCA & ACC\\ \hline
LSTM$-2 \times 64-l2$ & 4.08 & 2.19, 3.03 & 0.78$\pm$6.32 & \textbf{1.00$\pm$0.06} & 0.95$\pm$0.04\\ \hline
LSTM$-2 \times 128-l2$ & 2.83 & 1.56, 2.05 & 0.50$\pm$4.73 & \textbf{1.00$\pm$0.05} & 0.96$\pm$0.03\\ \hline
LSTM$-2 \times 256-l2$ & 5.87 & 2.83, 4.70 & 2.00$\pm$8.97 & 0.98$\pm$0.09 & 0.93$\pm$0.06\\ \hline
LSTM$-2 \times 128-$xyz & 5.74 & 3.02, 4.32 & 1.56$\pm$8.69 & 0.98$\pm$0.09 & 0.93$\pm$0.05\\ \hline
LSTM$-2 \times 128-l2$ and xyz & 4.42 & 2.15, 3.52 & 1.45$\pm$7.13 & 0.98$\pm$0.07 & 0.94$\pm$0.05\\ \hline
\hline
LSTM$-2 \times 64-l2$ (attention) & 5.43 & 3.03, 3.94 & 1.12$\pm$8.34 & 0.99$\pm$0.08 & 0.93$\pm$0.05\\ \hline
LSTM$-2 \times 128-l2$ (attention) & \textbf{2.33} & \textbf{1.39}, 1.60 & \textbf{0.17$\pm$3.92} & \textbf{1.00$\pm$0.04} & \textbf{0.97$\pm$0.03}\\ \hline
LSTM$-2 \times 256-l2$ (attention) & 5.38 & 2.89, 4.01 & 1.29$\pm$8.1 & 0.99$\pm$0.08 & 0.93$\pm$0.05\\ \hline
LSTM$-2 \times 128-$xyz (attention) & 4.03 & 2.27, 2.90 & 0.61$\pm$6.71 & 0.99$\pm$0.07 & 0.95$\pm$0.04\\ \hline
LSTM$-2 \times 128-l2$ and xyz (attention) & 2.69 & 1.95, \textbf{1.50} & $-0.55\pm$4.80 & \textbf{1.00$\pm$0.05} & 0.96$\pm$0.03\\ \hline
\end{tabular}
\label{tab:wdsc}
\end{table*}

\begin{table*}[ht]
\caption{The results of the proposed SC method on different populations of the WeAllWalk dataset \cite{flores2018} using the LSTM with and without attention with 2 layers of 128 neurons, and the $l2$-norm of the accelerometer signal (see Section \ref{sec:experimental_settings}) using different evaluation metrics (see Section \ref{sec:evaluation_metrics}).}
\centering
\begin{tabular}{
p{.3\linewidth}p{.075\linewidth}p{.075\linewidth}p{.085\linewidth}p{.085\linewidth}p{.085\linewidth}
}
\hline
 & MAE & UC, OC & ER & RCA & ACC\\ \hline
Sighted Subjects \\ \hline
LSTM$-2 \times 128-$l2 & 2.51 & 2.80, 4.03 & \textbf{3.67$\pm$10.76} & \textbf{0.97$\pm$0.1} & 0.92$\pm$0.07\\ \hline
LSTM$-2 \times 128-$l2 (attention) & \textbf{1.25} & \textbf{2.20, 1.46} & -5.83$\pm$12.88 & 1.05$\pm$0.12 & \textbf{0.93$\pm$0.11}\\ \hline
\hline
Blind Subject with a White Cane \\ \hline
LSTM$-2 \times 128-$l2 & 6.80 & 4.93, 9.49 & 7.6$\pm$23.32 & 0.92$\pm$0.24 & 0.83$\pm$0.18\\ \hline
LSTM$-2 \times 128-$l2 (attention) & \textbf{4.09} & \textbf{3.50, 5.10} & \textbf{-2.36$\pm$13.35} & \textbf{1.02$\pm$0.14} & \textbf{0.89$\pm$0.09} \\ \hline
\hline
Blind Subject with a Guide Dog \\ \hline
LSTM$-2 \times 128-$l2 & 4.64 & 5.53, 5.91 & \textbf{4.16$\pm$26.25} & \textbf{0.96$\pm$0.27} & 0.84$\pm$0.22\\ \hline
LSTM$-2 \times 128-$l2 (attention) & \textbf{2.51} & \textbf{3.64, 2.56} & -8.47$\pm$18.68 & 1.08$\pm$0.19 & \textbf{0.88$\pm$0.17}\\ \hline
\end{tabular}
\label{tab:weallwalk}
\end{table*}

\begin{table*}[ht]
\caption{The results of the proposed SC method on different walking regularities and sensor placements in the Pedometer dataset \cite{mattfeld2017} using the LSTM with and without attention with 2 layers of 128 neurons, and the $l2$-norm of the accelerometer signal (see Section \ref{sec:experimental_settings}) using different evaluation metrics (see Section \ref{sec:evaluation_metrics}).}
\centering
\begin{tabular}{
p{.3\linewidth}p{.075\linewidth}p{.085\linewidth}p{.085\linewidth}p{.085\linewidth}p{.085\linewidth}
}
\hline
 & MAE & UC, OC & ER & RCA & ACC\\ \hline

Regular Walking$-$Sensor on Ankle \\ \hline
LSTM$-2 \times 128-$l2 & 5.21 & 1.71, 1.87 & -0.17$\pm$6.53 & 1.00$\pm$0.06 & 0.96$\pm$0.05\\ \hline
LSTM$-2 \times 128-$l2 (attention) & \textbf{2.00} & \textbf{1.04, 0.57} & \textbf{-0.68$\pm$3.77} & \textbf{1.00$\pm$0.037} & \textbf{0.98$\pm$0.03}\\ \hline
\hline

Semi-regular Walking$-$Sensor on Ankle \\ \hline
LSTM$-2 \times 128-$l2 & \textbf{13.87} & 7.21, 8.39 & -2.84$\pm$21.93 & 1.03$\pm$0.22 & 0.83$\pm$0.14\\ \hline
LSTM$-2 \times 128-$l2 (attention) & 19.67 & \textbf{5.44, 5.68} & \textbf{-1.75$\pm$14.69} & \textbf{1.02$\pm$0.15} & \textbf{0.88$\pm$0.09}\\ \hline
\hline

Regular Walking$-$Sensor on Wrist \\ \hline
LSTM$-2 \times 128-$l2 & 5.65 & 2.20, 2.36 & \textbf{-0.22$\pm$6.50} & \textbf{1.00$\pm$0.06} & 0.95$\pm$0.05\\ \hline
LSTM$-2 \times 128-$l2 (attention) & \textbf{2.40} & \textbf{1.53, 0.40} & -1.54$\pm$7.70 & 1.02$\pm$0.080 & \textbf{0.98$\pm$0.07}\\ \hline
\hline

Semi-regular Walking$-$Sensor on Wrist \\ \hline
LSTM$-2 \times 128-$l2 & \textbf{13.95} & 9.26, 6.81 & -7.46$\pm$25.43 & 1.07$\pm$0.25 & 0.81$\pm$0.19\\ \hline
LSTM$-2 \times 128-$l2 (attention) & 14.59 & \textbf{7.69, 8.82} & \textbf{-3.4$\pm$24.73} & \textbf{1.03$\pm$0.25} & \textbf{0.82$\pm$0.17}\\ \hline
\hline

Regular Walking$-$Sensor on Hip \\ \hline
LSTM$-2 \times 128-$l2 & 5.37 & 2.11, 2.22 & -0.26$\pm$6.40 & 1.00$\pm$0.07 & 0.96$\pm$0.05\\ \hline
LSTM$-2 \times 128-$l2 (attention) & \textbf{1.27} & \textbf{0.53, 0.50} & \textbf{-0.04$\pm$1.89} & \textbf{1.00$\pm$0.02} & \textbf{0.99$\pm$0.02}\\ \hline
\hline

Semi-regular Walking$-$Sensor on Hip \\ \hline
LSTM$-2 \times 128-$l2 & \textbf{9.29} & 10.39, \textbf{10.68} & -7.72$\pm$35.81 & 1.07$\pm$0.35 & 0.75$\pm$0.27\\ \hline
LSTM$-2 \times 128-$l2 (attention) & 9.38 & \textbf{7.56}, 13.35 & \textbf{-1.29$\pm$31.33} & \textbf{1.00$\pm$0.32} & \textbf{0.77$\pm$0.21}\\ \hline

\hline
\end{tabular}
\label{tab:pedomter}
\end{table*}

\subsection{Experimental Results}
\label{sec:experimental_results}
The proposed method is evaluated using various evaluation metrics described in Section \ref{sec:evaluation_metrics}. On the WDSC dataset \cite{brajdic2013}, five-fold cross-validation is performed with the same network architecture across all folds. The average (and standard deviation) results of all samples across all folds using different evaluation metrics are presented in Table \ref{tab:wdsc}. The results are shown for the LSTM without attention and the attention-based LSTM (described in Section \ref{sec:experimental_settings}) with different numbers of neurons in the hidden layers of the LSTM and correspondingly its following linear layers (including the attention layer), and different signal dimensionality, the raw x, y, and z components of accelerometer signals (3-dimensional), the $l2$-norm of the x, y, and z components (1-dimensional), and both (4-dimensional). According to Table \ref{tab:wdsc}, in all the configurations of LSTM and attention-based LSTM, a 2-layer LSTM with 128 neurons in the hidden layer has the best performance. The $l2$-norm itself works better than the raw components and both the $l2$-norm and raw components. In almost all the configurations of the vanilla LSTM, and attention-based LSTM, in the upper, and lower halves of Table \ref{tab:wdsc}, respectively, adding attention mechanism to the vanilla LSTM results in significant improvements in different evaluation metrics. The best attention-based model achieves (in bold letters) very low values of MAE, UC and OC (Flores \cite{flores2018}), and ER (Brajdic \cite{brajdic2013}), and very close to one values of RCA (Mattfeld \cite{mattfeld2021}) and very high values of ACC (Luu \cite{luu2022}).

Fig. \ref{fig:exemplary} (a) illustrate  the raw x, y, and z components, \ref{fig:exemplary} (b) the normalized $l2$-norm of (a), and \ref{fig:exemplary} (c) the energy (corresponding to the signal in \ref{fig:exemplary} (b)) of the attention mechanism in a trained attention-based LSTM neural network (the best model in Table \ref{tab:wdsc}), for the first half of an exemplary accelerometer signal in the WDSC dataset \cite{brajdic2013}. As shown in Fig. \ref{fig:exemplary} (c), the energy, as the output of the attention layer, has a shape in which the steps have been emphasized. As described in Section \ref{sec:method}, the energy and its corresponding weights will be multiplied by the output of the LSTM in the network. In this way, the attention mechanism learns and pays attention to the steps, modifies the output of the LSTM accordingly, and the neural network outputs the step count.

Table \ref{tab:weallwalk} presents the results of the proposed SC method using the LSTM without attention and the attention-based LSTM with 128 neurons in the two hidden layers of the LSTM and the $l2$-norm of the accelerometer signal using different evaluation metrics for the WeAllWalk dataset \cite{flores2018}. Following the experimental settings in the original work introduced in the WeAllWalk dataset \cite{flores2018}, leave-one-person-out cross-validation in different populations is implemented. The results are presented for sighted subjects, blind subjects with a white cane, and blind subjects with a guide dog to examine the robustness of the proposed method in different populations with different levels of walking regularity. As can be seen in Table \ref{tab:weallwalk}, in almost all the populations, in most of the evaluation metrics, adding attention significantly improves the SC performance. The proposed attention-based method is robust against irregular walking, i.e., in blind subjects with a white cane and blind subjects with a guide dog.

Table \ref{tab:pedomter} presents the results of the proposed SC method using the LSTM without attention and the attention-based LSTM with 128 neurons in the two hidden layers of LSTM and the $l2$-norm of the accelerometer signal using different evaluation metrics (see Section \ref{sec:evaluation_metrics}) for the Pedometer dataset \cite{mattfeld2017}. Following the experimental settings in the original work introduced the Pedometer dataset \cite{mattfeld2021}, leave-two-person-out cross-validation in two levels of walking regularity and three different sensor placements are implemented, and the average results are reported in six sections of Table \ref{tab:pedomter}. As can be seen in Table \ref{tab:pedomter}, in most of the sections using most of the evaluation metrics, adding attention improves the SC performance. However, compared to the LSTM, the performance deterioration (from regular to semi-regular walking) is more severe in the attention-based LSTM.

Table \ref{tab:pedomter_comparison_regular} shows the results of the proposed LSTM and attention-based LSTM SC methods with the hyper-parameters in the previous experiment on the Pedometer dataset \cite{mattfeld2017}, compared to the previous deep-learning-based methods on the whole regular walking samples in the Pedometer dataset \cite{mattfeld2017} using leave-two-person-out cross-validation. Our proposed attention-based LSTM method competes with the CNN method \cite{luu2022}. As explained in Section \ref{sec:related_work}, the CNN method \cite{luu2022} is based on windowing and requires determining the window size. In addition, contrary to the proposed method, which only requires one single annotation data (the number of steps) for the entire signal, the CNN method \cite{luu2022} requires step annotation data at each timestep of the signal.

Table \ref{tab:pedomter_comparison_regular_semiregular} shows the results of the proposed LSTM and attention-based LSTM SC methods with the hyper-parameters in the previous experiment on the Pedometer dataset \cite{mattfeld2017} compared to the previous time-domain methods \cite{mattfeld2021} on the whole regular walking samples and semi-regular walking samples in the Pedometer dataset \cite{mattfeld2017} using leave-two-person-out cross validation. Our proposed attention-based LSTM method outperforms the previous time-domain methods \cite{mattfeld2021} with the advantage of not requiring many parameters and thresholds as in the time domain methods.

\begin{table}
\caption{The ACC \cite{luu2022} of the proposed LSTM and attention-based LSTM SC method compared to the previous deep-learning methods \cite{luu2022} on the whole (all the three sensor placements) regular walking samples in the Pedometer dataset
\cite{mattfeld2017} (see Sections IV-C and IV-B).}
\centering
\begin{tabular}{
p{.35\linewidth}p{.15\linewidth}
}
\hline
 & ACC\\ \hline
LSTM \cite{luu2022} & 0.9487\\ \hline
WaveNet \cite{luu2022} & 0.9851\\ \hline
CNN \cite{luu2022} & \textbf{0.9872}\\ \hline

LSTM$-2 \times 128-$l2 & 0.9513\\ \hline
LSTM$-2 \times 128-$l2 (att) & \textbf{0.9868}\\ \hline
\end{tabular}
\label{tab:pedomter_comparison_regular}
\end{table}

\begin{table}
\caption{The mean and standard deviation of the RCA \cite{mattfeld2021} for the proposed LSTM with and without attention with 2 layers of 128 neurons compared to the previous time-domain methods \cite{mattfeld2021} on the whole (all the three sensor placements) regular and semi-regular walking samples in the Pedometer dataset \cite{mattfeld2017} (see Sections \ref{sec:experimental_settings} and \ref{sec:evaluation_metrics}).
}
\centering
\begin{tabular}{
p{.35\linewidth}p{.3\linewidth}p{.2\linewidth}
}
\hline
 & \multicolumn{2}{c}{RCA}\\ \hline
 & Regular & Semi-regular\\ \hline
Peak \cite{mattfeld2021} & 0.92$\pm$0.11 & 1.30$\pm$0.21\\ \hline
Threshold \cite{mattfeld2021} & 1.03$\pm$0.17 & 1.34$\pm$0.17\\ \hline
Autocorrelation \cite{mattfeld2021} & 0.95$\pm$0.24 & 0.93$\pm$0.17\\ \hline

LSTM$-2 \times 128-$l2 & \textbf{1.00$\pm$0.04} & 1.07$\pm$0.29\\ \hline
LSTM$-2 \times 128-$l2 (att) & \textbf{1.00$\pm$0.06} & \textbf{1.03$\pm$0.21}\\ \hline
\end{tabular}
\label{tab:pedomter_comparison_regular_semiregular}
\end{table}

\section{Conclusions}
\label{sec:conclusions}
This paper defined SC as a regression problem and used a many-to-one attention-based LSTM to solve it. Most of the previous methods work on windowed accelerometer signals in which SC results from step detection in individual signal windows. Our proposed method, working at the signal level, analyzes the entire accelerometer signal as a whole and outputs the number of steps. This signal level analysis eliminates the need for determining the window size and having ground-truth labels for every individual step. The proposed attention mechanism for RNNs that is capable of analyzing variable-length time series (signals) learns to pay attention to the steps and outputs their summation. The internal step identification is a consequence of applying the attention mechanism that is learned through the training of the neural network. The experimental results on three publicly available SC datasets demonstrated that the proposed method successfully counts the number of steps with low values of mean absolute error and high values of SC accuracy. Temporal Convolution Networks (TCNs) \cite{bai2018} are powerful neural networks for the analysis and modeling of sequences with extensive lengths. However, they are unable to handle variable-length sequences. Our future work will investigate modifying TCNs \cite{bai2018} and attention-based TCNs \cite{hao2020} to handle variable-length sequences and applying them to signal-level SC. In addition, we plan to work on developing personalized SC models for specific populations and individuals.



\end{document}